\documentclass[11pt,a4paper]{article}
\usepackage[hyperref]{acl2020}

\usepackage{times}
\usepackage{latexsym}
\usepackage{caption}
\usepackage{subcaption}
\usepackage{graphicx}
\usepackage{booktabs}
\usepackage{siunitx}

\usepackage{amsmath}
\usepackage{adjustbox}
\usepackage{arydshln}
\usepackage{hyperref}
\usepackage{microtype}
\usepackage{multirow}
\usepackage{scalerel}
\usepackage{longtable}

\usepackage[utf8]{inputenc}

\graphicspath{ {./Figures/} }

\usepackage{float}
\newfloat{footnote1}{hb}

\newcommand{\reffig}[1]{Figure \ref{#1}}
\newcommand{\reftbl}[1]{Table \ref{#1}}
\newcommand{\refsec}[1]{Section \ref{#1}}
\newcommand{\refapp}[1]{Appendix \ref{#1}}

\aclfinalcopy 

\newcommand\blfootnote[1]{%
  \begingroup
  \renewcommand\thefootnote{}\footnote{#1}%
  \addtocounter{footnote}{-1}%
  \endgroup
}

\title{MuRIL: Multilingual Representations for Indian Languages}

\author{
\vspace{2mm}
Simran Khanuja\textsuperscript{1} 
\quad Diksha Bansal\textsuperscript{* 2} 
\quad \textbf{Sarvesh Mehtani\textsuperscript{* 3}} 
\quad \textbf{Savya Khosla\textsuperscript{* 4}} 
\quad \textbf{Atreyee Dey\textsuperscript{1}} \\
\vspace{2mm}
\quad \textbf{Balaji Gopalan\textsuperscript{1}} 
\quad \textbf{Dilip Kumar Margam\textsuperscript{1}} 
\quad \textbf{Pooja Aggarwal\textsuperscript{1}}
\quad \textbf{Rajiv Teja Nagipogu\textsuperscript{1}}
\quad \textbf{Shachi Dave\textsuperscript{1}} \\
\vspace{4mm}
\quad \textbf{Shruti Gupta\textsuperscript{1}} 
\quad \textbf{Subhash Chandra Bose Gali\textsuperscript{1}} 
\quad \textbf{Vish Subramanian\textsuperscript{1}} 
\quad \textbf{Partha Talukdar\textsuperscript{1}}\\
\vspace{2mm}
\textsuperscript{1}Google 
\quad \textsuperscript{2}Indian Institute of Technology, Patna \\
\textsuperscript{3}Indian Institute of Technology, Bombay
\quad \textsuperscript{4}Delhi Technological University \\ }

\date{}

\begin{document}
\maketitle


\section{Why MuRIL?}
\begin{figure}
\begin{center}
\includegraphics[scale=0.31]{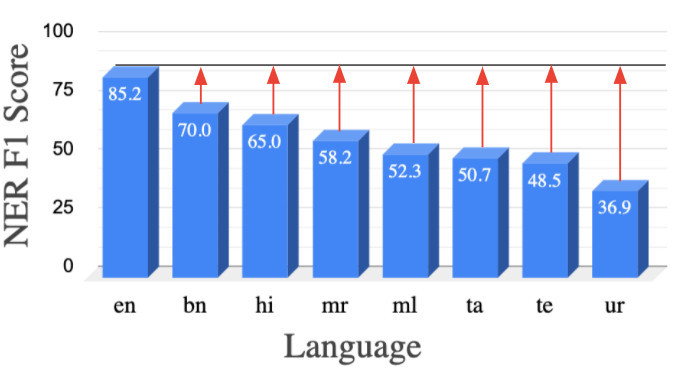} 
\caption{\emph{mBERT's (zero-shot) performance on Named Entity Recognition (NER)}. We observe significant differences between the performance on the English test set and other IN languages. This pattern is representative of current state-of-the-art multilingual models for Indian (IN) languages.}
\label{fig:ner_mbert}
\end{center}
\end{figure}

India is a multilingual society with 1369 rationalized languages and dialects being spoken across the country \cite{census}\blfootnote{* Work done during a summer internship at Google India. Correspondence to the MuRIL Team \texttt{(muril-contact@google.com)}}. Of these, the 22 scheduled languages have a staggering total of 1.17 billion speakers and 121 languages have more than 10,000 speakers \cite{census}. India also has the second largest (and an ever growing) digital footprint \cite{statista}. Despite this, today's state-of-the-art multilingual systems perform sub-optimally on Indian (IN) languages (as shown in Figure \ref{fig:ner_mbert}). This can be explained by the fact that multilingual language models (LMs) are often trained on 100+ languages together, leading to a small representation of IN languages in their vocabulary and training data. Multilingual LMs are substantially less effective in resource-lean scenarios \cite{wu-dredze-2020-languages, lauscher-etal-2020-zero}, as limited data doesn't help capture the various nuances of a language. One also commonly observes IN language text transliterated to Latin or code-mixed with English, especially in informal settings (for example, on social media platforms) \cite{rijhwani-etal-2017-estimating}. This phenomenon is not adequately handled by current state-of-the-art multilingual LMs. Few works like \citet{conneau-etal-2020-unsupervised} use transliterated data in training, but limit to including naturally present web crawl data. 

To address the aforementioned gaps, we propose MuRIL, a multilingual LM specifically built for IN languages. MuRIL is trained on significantly large amounts of IN text corpora \emph{only}. We explicitly augment monolingual text corpora with \emph{both} translated and transliterated document pairs, that serve as supervised cross-lingual signals in training. MuRIL significantly outperforms multilingual BERT (mBERT) on all tasks in the challenging cross-lingual XTREME benchmark \cite{hu2020xtreme}. We also present results on transliterated (native $\mathrm{\rightarrow}$ Latin) test sets of the chosen datasets, and demonstrate the efficacy of MuRIL in handling transliterated data. 

\section{Model and Data}
\label{model_data}
MuRIL currently supports 17 languages for which monolingual data is publicly available. These are further grouped into 16 IN languages and English (\emph{en}). The IN languages include: Assamese (\emph{as}), Bengali (\emph{bn}), Gujarati (\emph{gu}), Hindi (\emph{hi}), Kannada (\emph{kn}), Kashmiri (\emph{ks}), Malayalam (\emph{ml}), Marathi (\emph{mr}), Nepali (\emph{ne}), Oriya (\emph{or}), Punjabi (\emph{pa}), Sanskrit (\emph{sa}), Sindhi (\emph{sd}), Tamil (\emph{ta}), Telugu (\emph{te}) and Urdu (\emph{ur}).\\

\noindent We train our model with two language modeling objectives. The first is the conventional \emph{Masked Language Modeling} (MLM) objective \cite{taylor1953cloze} that leverages monolingual text data only (unsupervised). The second is the \emph{Translation Language Modeling} (TLM) objective \cite{lample2019cross} that leverages parallel data (supervised). We use monolingual documents to train the model with MLM, and \emph{both} translated and transliterated document pairs to train the model with TLM. \\

\noindent \textbf{Monolingual Data}: We collect monolingual data for the 17 languages mentioned above from the Common Crawl OSCAR corpus\footnote{\url{https://oscar-corpus.com}} and Wikipedia\footnote{\url{https://www.tensorflow.org/datasets/catalog/wikipedia}}. \\

\begin{figure}
\begin{center}
\includegraphics[scale=0.17]{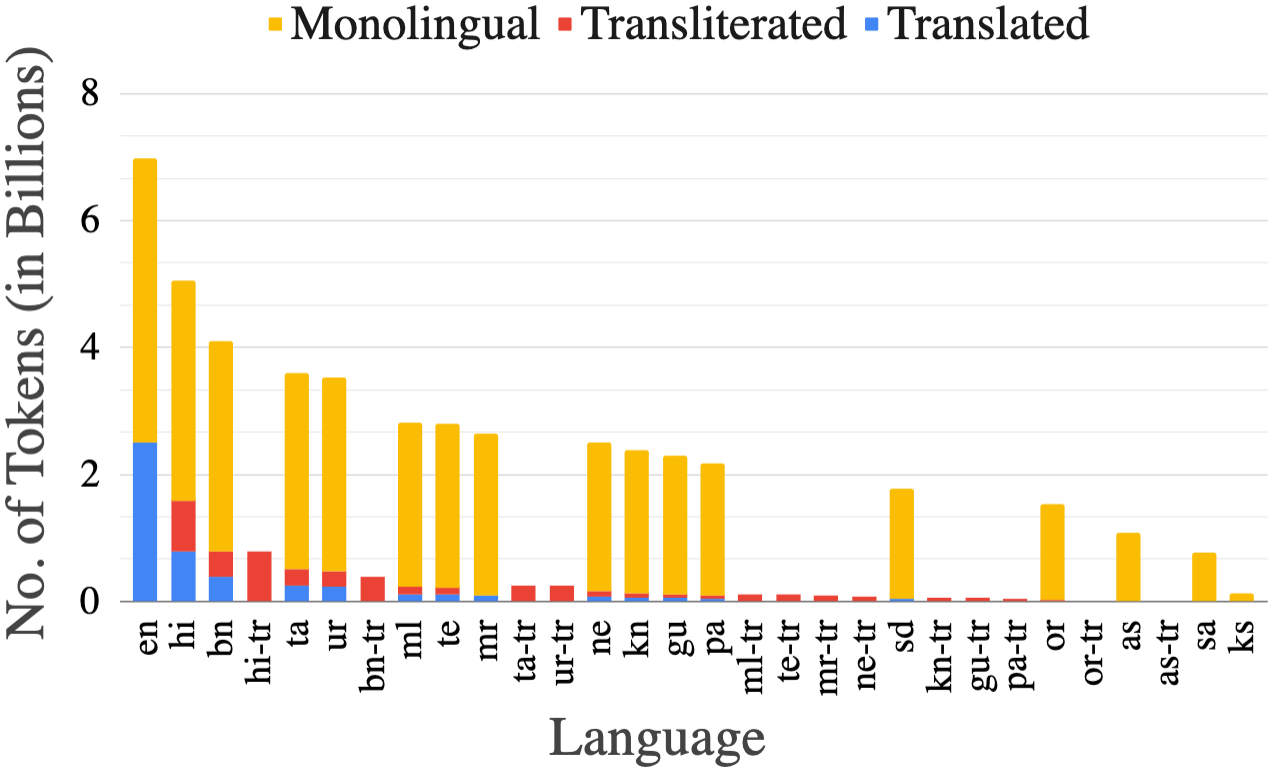} 
\caption{\emph{Upsampled Token Distribution}. We upsample monolingual Wikipedia corpora as described in \refsec{model_data}, to enhance low resource language representation in the pre-training data.}
\label{fig:upsampled_data}
\end{center}
\end{figure}

\noindent \textbf{Translated Data}: We have two sources of translated data. First, we use the PMINDIA \cite{haddow2020pmindia} parallel corpus containing sentence pairs for 8 IN languages (\emph{bn, gu, hi, kn, ml, mr, ta, te}). Each pair comprises of a sentence in a native language and its English translation. Second, we translate the aforementioned monolingual corpora (both Common Crawl and Wikipedia) to English, using an in-house translation system. The source and translated documents are used as parallel instances to train the model. Note that we translate corpora of all IN languages excluding \emph{as}, \emph{ks} and \emph{sa}, for which the current translation system lacks support. \\

\begin{figure}[t]
\begin{center}
\includegraphics[scale=0.22]{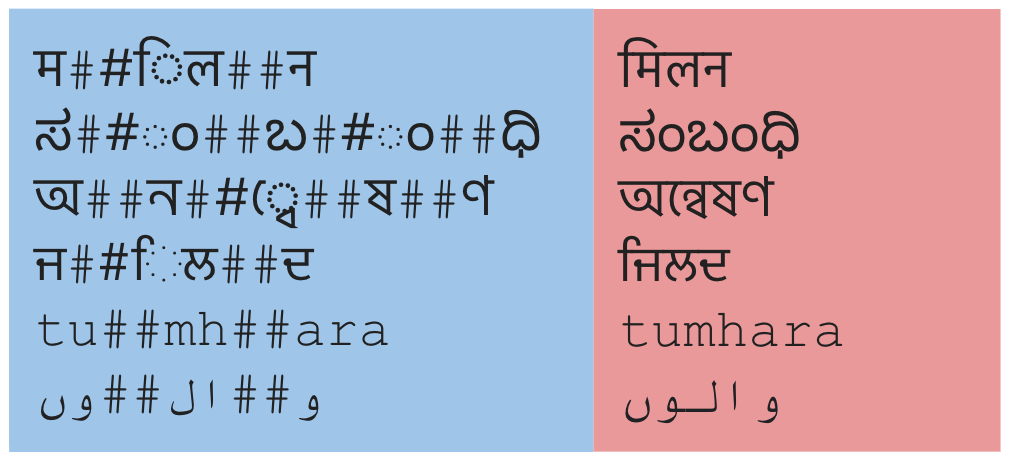} 
\caption{\emph{IN language words tokenized using mBERT (blue) and MuRIL (Red).}}
\label{fig:words_tokenize}
\end{center}
\end{figure}

\noindent \textbf{Transliterated Data}: We have two sources of transliterated data as well. First, we use the Dakshina Dataset \cite{roark-etal-2020-processing} that contains 10,000 sentence pairs for 12 IN languages (\emph{bn, gu, hi, kn, ml, mr, pa, ta, te, ur}). Each pair is a native script sentence and its manually romanized transliteration. Second, we use the \emph{indic-trans} library \cite{Bhat:2014:ISS:2824864.2824872} to transliterate Wikipedia corpora of all IN languages to Latin (except \emph{ks}, \emph{sa} and \emph{sd}, for which the library doesn't have  support). The source document and its Latin transliteration are used as parallel instances to train the model. \\

\noindent \textbf{Upsampling}: In the corpora collected above, the percentage of tokens per language is highly uneven in its distribution. Hence, data smoothing is essential so that all languages have their representation reflect their usage in the real world. To achieve this, we upsample monolingual Wikipedia corpora of each language according to its multiplier value given by:  
\begin{equation}
    m_i = \left(\frac{\max\limits_{j\in L}{n_j}}{n_i}\right)^{(1-\alpha)}
\end{equation}
In the above equation, $\mathrm{m_i}$ represents the multiplier value for language $\mathrm{i}$, $\mathrm{n_i}$ is its original token count, $\mathrm{L}$ represents the set of all 17 languages and $\mathrm{\alpha}$ is a hyperparameter whose value is set to 0.3, following  \citet{conneau-etal-2020-unsupervised}. Hence, the upsampled token count for language $\mathrm{i}$ is $\mathrm{m_i*n_i}$. The final data distribution after upsampling is shown in Figure \ref{fig:upsampled_data}. The upsampled token counts for each language and corpus are reported in \refapp{pretraining_data}. \\

\begin{figure}[t]
\begin{center}
\includegraphics[scale=0.165]{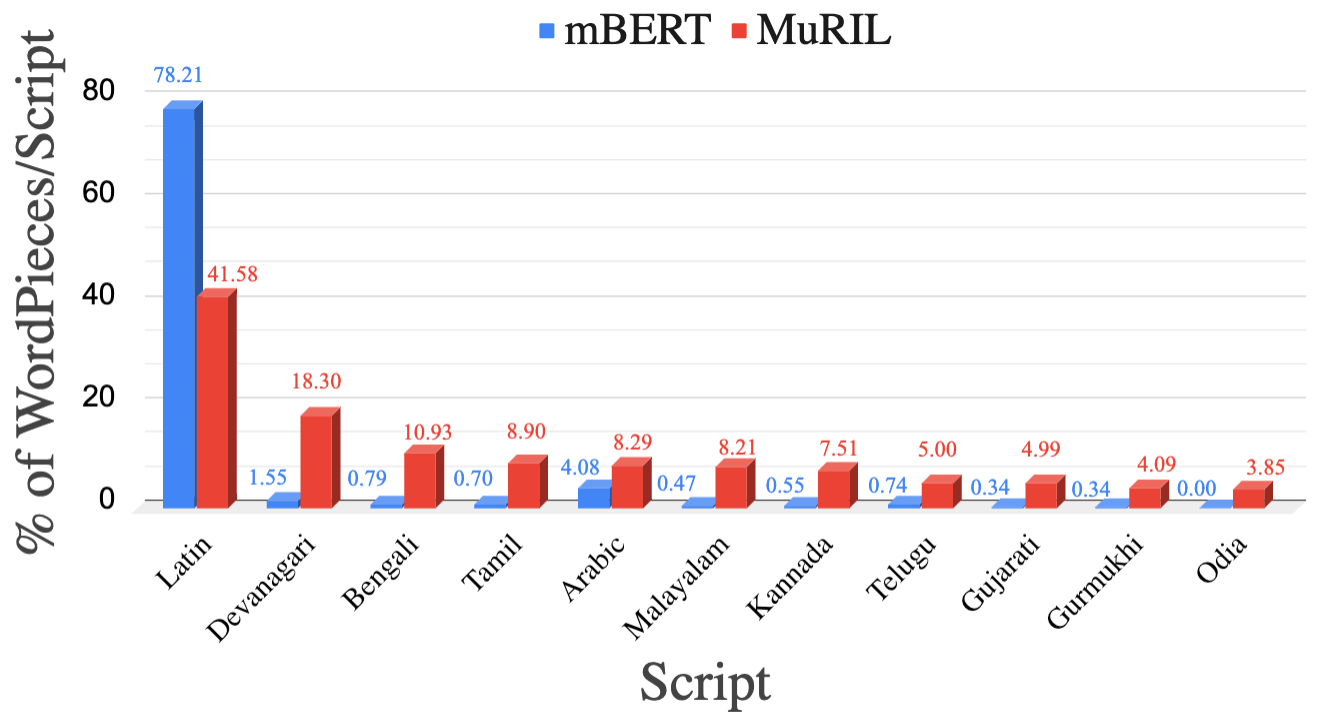} 
\caption{\emph{Percentage of WordPieces/Script} in mBERT and MuRIL vocabularies. A WordPiece belongs to the category if all of its characters fall into the category or are digits.}
\label{fig:mbert_muril_vocab}
\end{center}
\end{figure}

\begin{figure}[t]
\begin{center}
\includegraphics[scale=0.23]{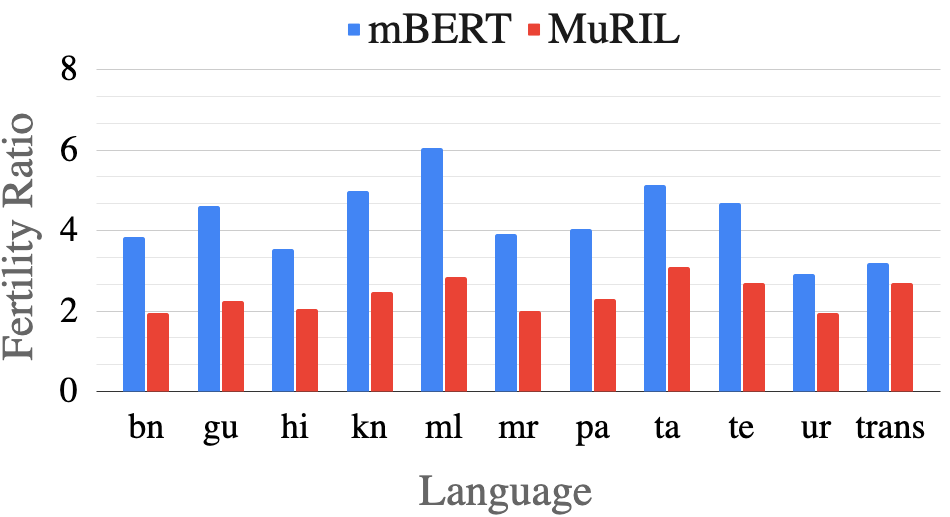} 
\caption{\emph{Fertility Ratio for IN languages using mBERT and MuRIL tokenizers.} Here, \emph{trans} subsumes all IN languages transliterated from their native script to Latin.}
\label{fig:fertility}
\end{center}
\end{figure}

\noindent \textbf{Vocabulary}: 
We learn a cased WordPiece \cite{schuster2012japanese, wu2016googles} vocabulary from the upsampled pre-training data using the wordpiece vocabulary generation library from Tensorflow Text\footnote{\url{https://github.com/tensorflow/text/blob/master/tensorflow\_text/tools/wordpiece\_vocab/generate\_vocab.py}}. Since our data is upsampled, we set the language smoothing exponent from the vocabulary generation tool to $\mathrm{1}$, and the rest of the parameters are set to their default value. The final vocabulary size is \textbf{197,285}. Figure \ref{fig:words_tokenize}  shows a few common IN language words tokenized using mBERT and MuRIL vocabularies. We also plot the \emph{fertility ratio} (average number of sub-words/word) of mBERT and MuRIL tokenizers on a random sample of text from our training data in \reffig{fig:fertility}. Here, a higher fertility ratio equates to a larger number of sub-words per word, eventually leading to a loss in preservation of semantic meaning. We observe a higher fertility ratio for mBERT as compared to MuRIL because of two reasons. First, there is very little representation of IN languages in the mBERT vocabulary\footnote{\url{http://juditacs.github.io/2019/02/19/bert-tokenization-stats.html}} (refer to Figure \ref{fig:mbert_muril_vocab} for a comparison) and second, the vocabulary does not take transliterated words into account. Since vocabulary plays a key role in the performance of transformer based LMs \cite{chung-etal-2020-improving, artetxe-etal-2020-cross}, MuRIL's vocabulary (specifically focused on IN languages) is a significant contributor to the model's improved performance over mBERT. \\

\begin{table*}[t]
\scalebox{1.0}{
\begin{tabular}{c | c c c c c c c | c}
\hline \multirow{2}{*}{Model} & {PANX} & {UDPOS} & {XNLI} & {Tatoeba} & {XQuAD} & {MLQA} & {TyDiQA-GoldP} & \multirow{2}{*}{\textbf{Avg.}} \\
\cdashline{2-8}
& \textbf{F1} & \textbf{F1} & \textbf{Acc.} &\textbf{Acc.} & \textbf{F1/EM} & \textbf{F1/EM} & \textbf{F1/EM} & \\
\hline
mBERT & 58.0 & 71.2 & 66.8 & 18.4 & 71.2/58.2 & 65.3/51.2 & 63.1/51.7 & 59.1 \\
MuRIL & \textbf{77.6} & \textbf{75.0} & \textbf{74.1} & \textbf{25.2} & \textbf{79.1/65.6} & \textbf{73.8/58.8} & \textbf{75.4/59.3} & \textbf{68.6} \\
\hline
\end{tabular}}
\caption{\label{MuRIL_results} \emph{Results for MuRIL and mBERT on XTREME (IN)}. We observe that MuRIL significantly outperforms mBERT on all the datasets in XTREME. Note that here we present the average performance on test sets for IN languages only that MuRIL currently supports. Please refer to \refsec{eval} for more details.}
\end{table*}

\begin{table*}[t]
\centering
\scalebox{1.0}{
\begin{tabular}{c | c c c c | c}
\hline \multirow{2}{*}{Model} & {PANX} & {UDPOS} & {XNLI} & {Tatoeba} & \multirow{2}{*}{\textbf{Avg.}} \\
\cdashline{2-5}
& \textbf{F1} & \textbf{F1} & \textbf{Acc.} &\textbf{Acc.} & \\
\hline
mBERT & 14.2 & 28.2 & 39.2 & 2.7 & 21.1 \\
MuRIL & \textbf{57.7} & \textbf{62.1} & \textbf{64.7} & \textbf{11.0} & \textbf{48.9} \\
\hline
\end{tabular}}
\caption{\label{transliterated_results} \emph{Results for MuRIL and mBERT on XTREME (IN-tr)}. We transliterate IN language test sets (native $\mathrm{\rightarrow}$ Latin) and present the average performance across all transliterated test sets. Please refer to \refsec{eval} for more details.}
\end{table*}

\begin{table*}[!htbp]
\centering
\scalebox{0.75}{
\begin{tabular}{c | c | c | c | c | c | c }
\hline \multirow{2}{*}{Task} & \multirow{2}{*}{Dataset} & \multirow{2}{*}{Batch Size} & \multirow{2}{*}{Learning Rate} & \multirow{2}{*}{No. of Epochs} & \multirow{2}{*}{Warmup Ratio} & Maximum Sequence \\ 
& & & & & & Length \\
\hline
\multirow{2}{*}{Structured Prediction} & PANX & 32 & 2e-5 & 10 & 0.1 & 128\\
& UDPOS & 32 & 2e-5 & 10 & 0.1 & 128 \\
\hdashline
Classification & XNLI & 32 & 2e-5 & 5 & 0.1 & 128 \\
\hdashline
\multirow{3}{*}{Question Answering} & XQuAD & 32 & 3e-5 & 2 & 0.1 & 384 \\
& MLQA & 32 & 3e-5 & 2 & 0.1 & 384 \\
& TyDiQA-GoldP & 32 & 3e-5 & 2 & 0.1 & 384 \\
\hline 
\end{tabular}}
\caption{\label{table:hyperparams}\emph{Fine-tuning hyperparameter details for each dataset in XTREME}. We use the same hyperparameters for fine-tuning both mBERT and MuRIL. Please refer to \refsec{eval} for more details.}
\end{table*}

\noindent \textbf{Pre-training Details}: We pre-train a BERT base encoder model making use of the MLM and TLM objectives. We keep a maximum sequence length of 512, a global batch size of 4096, and train for 1M steps (with 50k warm-up steps and a linear decay after). We make use of the AdamW optimizer with a learning rate of 5e-4. Our final model has 236M parameters, is trained on $\sim$16B unique tokens, and has a vocabulary of 197,285. Please note that we preserve the case to prevent stripping off accents, often present in IN languages.

\section{Evaluation}
\label{eval}
In all our experiments, the goal has been to improve the model’s performance for cross-lingual understanding. For this reason, the results are computed in a zero-shot setting, i.e., by fine-tuning models on the labeled training set of one language and evaluating on test sets for all languages. Here, our labeled training sets are in English for all tasks. We choose the XTREME benchmark \cite{hu2020xtreme} as a test-bed. XTREME covers 40 typologically diverse languages spanning 12 language families and includes 9 tasks that require reasoning about different levels of syntax or semantics \cite{hu2020xtreme}.

We present our results in \reftbl{MuRIL_results}. Since MuRIL currently supports IN languages only, we compute average performances across IN language test sets for all tasks. We also transliterate IN language test sets (native $\mathrm{\rightarrow}$ Latin) using the \emph{indic-trans} library \cite{Bhat:2014:ISS:2824864.2824872}, and report results on the same in \reftbl{transliterated_results}. Detailed results for each language and task can be found in \refapp{detailed_results}. On average, MuRIL significantly beats mBERT across \emph{all} tasks. This difference is more so for the transliterated test sets, which is expected because mBERT does not include transliterated data in training. We analyse predictions of mBERT and MuRIL on a random sample of test examples in \refapp{analysis}. \\

\noindent \textbf{Fine-tuning Details}: For each task, we report results of the best performing checkpoint on the evaluation set. We present the hyperparameter details for each task in Table \ref{table:hyperparams}. Note that we use the same hyperparameters for evaluating both mBERT and MuRIL. We fine-tune the model on the English training set for each task, and evaluate on the test sets of all IN languages. For TyDiQA-GoldP, we augment the training set with SQuAD English training set, similar to \citet{fang2020filter}, and then fine-tune the model. For Tatoeba, we do not fine-tune the model, and use the \emph{pooled\_output} of the last layer as the sentence embedding.

\section{How to use MuRIL?}
We have released the MuRIL encoder on TFHub\footnote{\url{https://tfhub.dev/google/MuRIL/1}} with detailed usage instructions. We have also released a pre-processor module with the same, that processes raw text into the expected input format for the encoder. Additionally, we have released the MuRIL pre-trained model, i.e., with the MLM layer intact (to enable masked word predictions) on HuggingFace\footnote{\url{https://huggingface.co/google/muril-base-cased}}. We sincerely hope MuRIL aids in building better technologies and applications for Indian languages.

\section*{Acknowledgments}
\noindent We would like to thank Melvin Johnson for his feedback on a draft of this paper. We would also like to thank Hyung Won Chung, Anosh Raj, Yinfei Yang and Fangxiaoyu Feng for contributing to our discussions around MuRIL. Finally, we would like to thank Nick Doiron for his feedback on the HuggingFace implementation of the model.
\bibliography{anthology,acl2020}

\begin{thebibliography}{17}
\expandafter\ifx\csname natexlab\endcsname\relax\def\natexlab#1{#1}\fi

\bibitem[{Artetxe et~al.(2020)Artetxe, Ruder, and
  Yogatama}]{artetxe-etal-2020-cross}
Mikel Artetxe, Sebastian Ruder, and Dani Yogatama. 2020.
\newblock \href {https://doi.org/10.18653/v1/2020.acl-main.421} {On the
  cross-lingual transferability of monolingual representations}.
\newblock In \emph{Proceedings of the 58th Annual Meeting of the Association
  for Computational Linguistics}, pages 4623--4637, Online. Association for
  Computational Linguistics.

\bibitem[{Bhat et~al.(2015)Bhat, Mujadia, Tammewar, Bhat, and
  Shrivastava}]{Bhat:2014:ISS:2824864.2824872}
Irshad~Ahmad Bhat, Vandan Mujadia, Aniruddha Tammewar, Riyaz~Ahmad Bhat, and
  Manish Shrivastava. 2015.
\newblock \href {https://doi.org/10.1145/2824864.2824872} {Iiit-h system
  submission for fire2014 shared task on transliterated search}.
\newblock In \emph{Proceedings of the Forum for Information Retrieval
  Evaluation}, FIRE '14, pages 48--53, New York, NY, USA. ACM.

\bibitem[{Chung et~al.(2020)Chung, Garrette, Tan, and
  Riesa}]{chung-etal-2020-improving}
Hyung~Won Chung, Dan Garrette, Kiat~Chuan Tan, and Jason Riesa. 2020.
\newblock \href {https://doi.org/10.18653/v1/2020.emnlp-main.367} {Improving
  multilingual models with language-clustered vocabularies}.
\newblock In \emph{Proceedings of the 2020 Conference on Empirical Methods in
  Natural Language Processing (EMNLP)}, pages 4536--4546, Online. Association
  for Computational Linguistics.

\bibitem[{Conneau et~al.(2020)Conneau, Khandelwal, Goyal, Chaudhary, Wenzek,
  Guzm{\'a}n, Grave, Ott, Zettlemoyer, and
  Stoyanov}]{conneau-etal-2020-unsupervised}
Alexis Conneau, Kartikay Khandelwal, Naman Goyal, Vishrav Chaudhary, Guillaume
  Wenzek, Francisco Guzm{\'a}n, Edouard Grave, Myle Ott, Luke Zettlemoyer, and
  Veselin Stoyanov. 2020.
\newblock \href {https://doi.org/10.18653/v1/2020.acl-main.747} {Unsupervised
  cross-lingual representation learning at scale}.
\newblock In \emph{Proceedings of the 58th Annual Meeting of the Association
  for Computational Linguistics}, pages 8440--8451, Online. Association for
  Computational Linguistics.

\bibitem[{Fang et~al.(2020)Fang, Wang, Gan, Sun, and Liu}]{fang2020filter}
Yuwei Fang, Shuohang Wang, Zhe Gan, Siqi Sun, and Jingjing Liu. 2020.
\newblock \href {https://arxiv.org/abs/2009.05166} {Filter: An enhanced fusion
  method for cross-lingual language understanding}.
\newblock \emph{arXiv preprint arXiv:2009.05166}.

\bibitem[{Haddow and Kirefu(2020)}]{haddow2020pmindia}
Barry Haddow and Faheem Kirefu. 2020.
\newblock \href {https://arxiv.org/abs/2001.09907} {Pmindia--a collection of
  parallel corpora of languages of india}.
\newblock \emph{arXiv preprint arXiv:2001.09907}.

\bibitem[{Hu et~al.(2020)Hu, Ruder, Siddhant, Neubig, Firat, and
  Johnson}]{hu2020xtreme}
Junjie Hu, Sebastian Ruder, Aditya Siddhant, Graham Neubig, Orhan Firat, and
  Melvin Johnson. 2020.
\newblock \href {https://arxiv.org/abs/2003.11080} {Xtreme: A massively
  multilingual multi-task benchmark for evaluating cross-lingual
  generalization}.
\newblock \emph{arXiv preprint arXiv:2003.11080}.

\bibitem[{INDIA(2011)}]{census}
INDIA. 2011.
\newblock Census of india, 2011.
\newblock
  \url{https://www.censusindia.gov.in/2011Census/C-16_25062018_NEW.pdf}.

\bibitem[{Lample and Conneau(2019)}]{lample2019cross}
Guillaume Lample and Alexis Conneau. 2019.
\newblock Cross-lingual language model pretraining.
\newblock \emph{arXiv preprint arXiv:1901.07291}.

\bibitem[{Lauscher et~al.(2020)Lauscher, Ravishankar, Vuli{\'c}, and
  Glava{\v{s}}}]{lauscher-etal-2020-zero}
Anne Lauscher, Vinit Ravishankar, Ivan Vuli{\'c}, and Goran Glava{\v{s}}. 2020.
\newblock \href {https://doi.org/10.18653/v1/2020.emnlp-main.363} {From zero to
  hero: {O}n the limitations of zero-shot language transfer with multilingual
  {T}ransformers}.
\newblock In \emph{Proceedings of the 2020 Conference on Empirical Methods in
  Natural Language Processing (EMNLP)}, pages 4483--4499, Online. Association
  for Computational Linguistics.

\bibitem[{Rijhwani et~al.(2017)Rijhwani, Sequiera, Choudhury, Bali, and
  Maddila}]{rijhwani-etal-2017-estimating}
Shruti Rijhwani, Royal Sequiera, Monojit Choudhury, Kalika Bali, and
  Chandra~Shekhar Maddila. 2017.
\newblock \href {https://doi.org/10.18653/v1/P17-1180} {Estimating
  code-switching on twitter with a novel generalized word-level language
  detection technique}.
\newblock In \emph{Proceedings of the 55th Annual Meeting of the Association
  for Computational Linguistics (Volume 1: Long Papers)}, pages 1971--1982,
  Vancouver, Canada. Association for Computational Linguistics.

\bibitem[{Roark et~al.(2020)Roark, Wolf-Sonkin, Kirov, Mielke, Johny,
  Demirsahin, and Hall}]{roark-etal-2020-processing}
Brian Roark, Lawrence Wolf-Sonkin, Christo Kirov, Sabrina~J. Mielke, Cibu
  Johny, Isin Demirsahin, and Keith Hall. 2020.
\newblock \href {https://www.aclweb.org/anthology/2020.lrec-1.294} {Processing
  {S}outh {A}sian languages written in the {L}atin script: the dakshina
  dataset}.
\newblock In \emph{Proceedings of the 12th Language Resources and Evaluation
  Conference}, pages 2413--2423, Marseille, France. European Language Resources
  Association.

\bibitem[{Schuster and Nakajima(2012)}]{schuster2012japanese}
Mike Schuster and Kaisuke Nakajima. 2012.
\newblock Japanese and korean voice search.
\newblock In \emph{2012 IEEE International Conference on Acoustics, Speech and
  Signal Processing (ICASSP)}, pages 5149--5152. IEEE.

\bibitem[{Statista(2020)}]{statista}
Statista. 2020.
\newblock Statista, 2020.
\newblock
  \url{https://www.statista.com/statistics/255146/number-of-internet-users-in-india/}.

\bibitem[{Taylor(1953)}]{taylor1953cloze}
Wilson~L Taylor. 1953.
\newblock \href {https://www.gwern.net/docs/psychology/writing/1953-taylor.pdf}
  {“cloze procedure”: A new tool for measuring readability}.
\newblock \emph{Journalism quarterly}, 30(4):415--433.

\bibitem[{Wu and Dredze(2020)}]{wu-dredze-2020-languages}
Shijie Wu and Mark Dredze. 2020.
\newblock \href {https://doi.org/10.18653/v1/2020.repl4nlp-1.16} {Are all
  languages created equal in multilingual {BERT}?}
\newblock In \emph{Proceedings of the 5th Workshop on Representation Learning
  for NLP}, pages 120--130, Online. Association for Computational Linguistics.

\bibitem[{Wu et~al.(2016)Wu, Schuster, Chen, Le, Norouzi, Macherey, Krikun,
  Cao, Gao, Macherey, Klingner, Shah, Johnson, Liu, Łukasz Kaiser, Gouws,
  Kato, Kudo, Kazawa, Stevens, Kurian, Patil, Wang, Young, Smith, Riesa,
  Rudnick, Vinyals, Corrado, Hughes, and Dean}]{wu2016googles}
Yonghui Wu, Mike Schuster, Zhifeng Chen, Quoc~V. Le, Mohammad Norouzi, Wolfgang
  Macherey, Maxim Krikun, Yuan Cao, Qin Gao, Klaus Macherey, Jeff Klingner,
  Apurva Shah, Melvin Johnson, Xiaobing Liu, Łukasz Kaiser, Stephan Gouws,
  Yoshikiyo Kato, Taku Kudo, Hideto Kazawa, Keith Stevens, George Kurian,
  Nishant Patil, Wei Wang, Cliff Young, Jason Smith, Jason Riesa, Alex Rudnick,
  Oriol Vinyals, Greg Corrado, Macduff Hughes, and Jeffrey Dean. 2016.
\newblock \href {http://arxiv.org/abs/1609.08144} {Google's neural machine
  translation system: Bridging the gap between human and machine translation}.

\end{thebibliography}
\bibliographystyle{acl_natbib}

\clearpage

\appendix

\begin{figure}
\begin{center}
\includegraphics[scale=0.37]{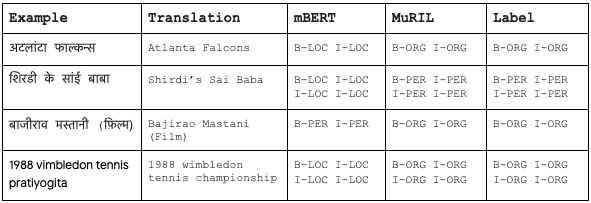} 
\caption{\emph{NER Predictions}}
\label{ner_analyse}
\end{center}
\end{figure}

\section{Pre-training Data Statistics}
\label{pretraining_data}
The upsampled token counts for each language and corpus are reported in \reftbl{table:data_stats}.

\section{Detailed Results}
\label{detailed_results}
We report per language results for each \emph{XTREME (IN)} dataset in Tables \ref{table:panx} (PANX), \ref{table:udpos} (UDPOS), \ref{table:xnli} (XNLI), \ref{table:tatoeba} (Tatoeba), \ref{table:qa} (XQuAD, MLQA) and \ref{table:tydiqa} (TyDiQA-GoldP). The detailed results for transliterated test sets are shown in Tables \ref{table:panx_tr} (PANX), \ref{table:udpos_tr} (UDPOS), \ref{table:xnli_tr} (XNLI), \ref{table:tatoeba_tr} (Tatoeba).


\section{Analysis}
\label{analysis}
In this section, we analyse the predictions of mBERT and MuRIL on a random sample of test examples.\\

\noindent \textbf{Named Entity Recognition (NER)}: NER is the task of locating and classifying entities in unstructured text into pre-defined categories such as person, location, organization etc. In \reffig{ner_analyse}, we present entity predictions of mBERT and MuRIL on a random sample of test examples.

In the first example, we observe that \emph{Atlanta Falcons}, a football team, is predicted as \emph{ORG} (Organisation) by MuRIL but \emph{LOC} (Location) by mBERT, probably looking at the word \emph{Atlanta} without context.

In the second example, MuRIL correctly takes the context into account and predicts \emph{Shirdi's Sai Baba} as \emph{PER} (Person), whereas mBERT resorts to predicting \emph{LOC} taking a cue from the word \emph{Shirdi}. A similar pattern is observed in other examples like \emph{Nepal's Prime Minister}, \emph{the President of America} etc.

In the third example, the \emph{Bajirao Mastani} movie is being spoken about, specified with the word (film) in parentheses, which MuRIL correctly captures. 

In the last example, we observe that MuRIL can correctly classify misspelled words (\emph{vimbledon}) utilising the context. \\

\begin{figure}
\begin{center}
\includegraphics[scale=0.28]{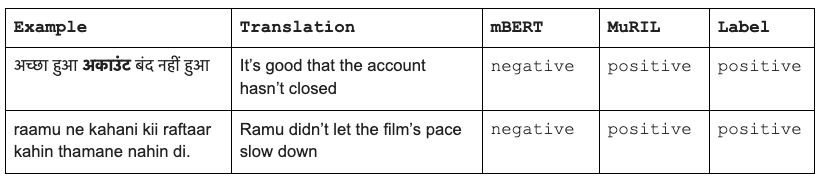} 
\caption{\emph{Sentiment Predictions}}
\label{sent_analyse}
\end{center}
\end{figure}

\noindent \textbf{Sentiment Analysis}: Sentiment analysis is a sentence classification task wherein each sentence is labeled to be expressing a positive, negative or neutral sentiment. We present the sentiment predictions on a sample set of sentences in \reffig{sent_analyse}. 

In the first example, \emph{``It’s good that the account hasn’t closed''}, we observe that the original Hindi sentence borrows an English word (\emph{account}) and also contains a negation (\emph{not}) word, but MuRIL correctly predicts it as expressing a positive statement. A similar observation can be made in the second example where MuRIL correctly predicts the sentiment of the transliterated sentence,  \emph{``Ramu didn't let the film's pace slow down''}. \\

\begin{figure}
\begin{center}
\includegraphics[scale=0.32]{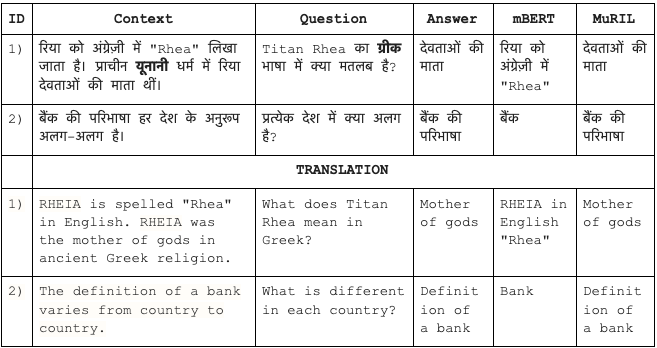} 
\caption{\emph{Question Answering}}
\label{qa_analyse}
\end{center}
\end{figure}

\noindent \textbf{Question Answering (QA)}: QA is the task of answering a question based on the given context or world knowledge. We show two context-question pairs, with their answers and predicted answers in \reffig{qa_analyse}.

In the first example, despite the fact that the word \emph{Greek} is referred to by its Hindi translation in the context and its transliteration in the question (as highlighted), MuRIL correctly infers the answer from the context.

In the second example, MuRIL understands that \emph{``bank ki paribhasha''} (the definition of a bank), as a whole entity, is what differs across countries and not banks.

\begin{table*}[t]
\centering
\scalebox{0.8}{
\begin{tabular}{c | c c | c c c | c c}
\hline \multirow{2}{*}{Language} & \multicolumn{2}{c|}{Monolingual} & \multicolumn{3}{c|}{Translated} &  \multicolumn{2}{c}{Transliterated} \\ 
\cline{2-8}
 & Wikipedia & Common Crawl & Wikipedia & Common Crawl & PMINDIA & Wikipedia & Dakshina \\
\hline
as & 2.5e+6 & 4.4e+6 & - & - & - & 2.5e+6 & - \\
as-tr & - & - & - & - & - & 2.5e+6 & - \\
bn & 2.7e+7 & 3.7e+8 & 2.7e+7 & 3.7e+8 & 4.4e+5 & 2.7e+7 & 1.2e+5 \\
bn-tr & - & - & - & - & - & 2.7e+7 & 1.2e+5 \\
en & 2.8e+9 & 1.7e+9 & 2.3e+8 & 2.3e+9 & 5.8e+6 & - & -\\
gu & 6.7e+6 & 5.1e+7 & 6.7e+6 & 5.1e+7 & 8.5e+5 & 6.7e+6 & 1.5e+5 \\
gu-tr & - & - & - & - & - & 6.7e+6 & 1.5e+5 \\
hi & 3.8e+7 & 7.5e+8 & 3.8e+7 & 7.5e+8 & 1.2e+6 & 3.8e+7 & 1.8e+5 \\
hi-tr & - & - & - & - & - & 3.8e+7 & 1.8e+5 \\
kn & 1.5e+7 & 5.0e+7 & 1.5e+7 & 5.0e+7 & 4.7e+5 & 1.5e+7 & 1.1e+5 \\
kn-tr & - & - & - & - & - & 1.5e+7 & 1.1e+5 \\
ks & 1.1e+4 & - & - & - & - & - & - \\
ml & 1.4e+7 & 9.8e+7 & 1.4e+7 & 9.8e+7 & 3.6e+5 & 1.4e+7 & 8.6e+4\\
ml-tr & - & - & - & - & - & 1.4e+7 & 8.6e+4 \\
mr & 8.3e+6 & 8.3e+7 & 8.3e+6 & 8.3e+7 & 5.2e+5 & 8.3e+6 & 9.8e+4 \\
mr-tr & - & - & - & - & - & 8.3e+6 & 9.8e+4 \\
ne & 5.0e+6 & 7.2e+7 & 5.0e+6 & 7.2e+7 & - & 5.0e+6 & - \\
ne-tr & - & - & - & - & - & 5.0e+6 & - \\
or & 3.4e+6 & 1.1e+7 & 3.4e+6 & 1.1e+7 & - & 3.4e+6 & -\\
or-tr & - & - & - & - & - & 3.4e+6 & - \\
pa & 9.1e+6 & 3.8e+7 & 9.1e+6 & 3.8e+7 & - & 9.1e+6 & 1.8e+5 \\
pa-tr & - & - & - & - & - & 9.1e+6 & 1.8e+5  \\
sa & 2.6e+6 & 1.7e+6 & - & - & - & - & - \\
sd & 3.4e+6 & 3.3e+7 & 3.4e+6 & 3.3e+7 & - & - & - \\
ta & 2.6e+7 & 2.3e+8 & 2.6e+7 & 2.3e+8 & 5.2e+5 & 2.6e+7 & 9.5e+4 \\
ta-tr & - & - & - & - & - & 2.6e+7 & 9.5e+4 \\
te & 3.0e+7 & 8.0e+7 & 3.0e+7 & 8.0e+7 & 5.7e+5 & 3.0e+7 & 9.4e+4 \\
te-tr & - & - & - & - & - & 3.0e+7 & 9.4e+4 \\
ur & 2.3e+7 & 2.2e+8 & 2.1e+7 & 2.2e+8 & - & 2.3e+7 & 1.7e+5 \\
ur-tr & - & - & - & - & - & 2.3e+7 & 1.7e+5 \\

\hline
\end{tabular}}
\caption{\label{table:data_stats} Number of tokens/corpus for each language. Note, X-tr stands for the transliterated counterpart of language X. }
\end{table*}

\begin{table*}[t]
\centering
\scalebox{1.0}{
\begin{tabular}{c | c c c c c c c c| c }
\hline Model & bn & en & hi & ml & mr & ta & te & ur & avg. \\ 
\hline 
mBERT & 68.6 & \textbf{84.4} & 65.1 & 54.8 & 58.4 & 51.2 & 50.2 & 31.4 & 58.0 \\
MuRIL & \textbf{86.0} & \textbf{84.4} & \textbf{78.1} & \textbf{75.8} & \textbf{74.6} & \textbf{71.9} & \textbf{65.0} & \textbf{85.1} & \textbf{77.6} \\
\hline
\end{tabular}}
\caption{\label{table:panx} PANX (F1) Results for each language. }
\end{table*}

\begin{table*}[t]
\centering
\scalebox{1.0}{
\begin{tabular}{c | c c c c c c | c }
\hline Model & en & hi & mr & ta & te & ur & avg. \\ 
\hline 
mBERT & 95.4 & \textbf{66.1} & 71.3 & 59.6 & 77.0 & 57.9 & 71.2 \\
MuRIL & \textbf{95.6} & 64.5 & \textbf{83.0} & \textbf{62.6} & \textbf{85.6} & \textbf{58.9} & \textbf{75.0} \\
\hline
\end{tabular}}
\caption{\label{table:udpos} UDPOS (F1) Results for each language. }
\end{table*}

\begin{table*}[t]
\centering
\scalebox{1.0}{
\begin{tabular}{c | c c c| c }
\hline Model & en & hi & ur & avg. \\ 
\hline 
mBERT & 81.7 & 60.5 & 58.2 & 66.8 \\
MuRIL & \textbf{83.9} & \textbf{70.7} & \textbf{67.7} & \textbf{74.1} \\
\hline
\end{tabular}}
\caption{\label{table:xnli} XNLI (Accuracy) Results for each language. }
\end{table*}

\begin{table*}[t]
\centering
\scalebox{1.0}{
\begin{tabular}{c | c c c c c c c | c }
\hline Model & bn & hi & ml & mr & ta & te & ur & avg. \\ 
\hline 
mBERT & 12.8 & 27.8 & 20.2 & 18.0 & 12.4 & 15.0 & 22.7 & 18.4 \\
MuRIL & \textbf{20.2} & \textbf{31.5} & \textbf{26.4} & \textbf{26.6} & \textbf{36.8} & \textbf{17.5} & \textbf{17.1} & \textbf{25.2} \\
\hline
\end{tabular}}
\caption{\label{table:tatoeba} Tatoeba (Accuracy) Results for each language. }
\end{table*}

\begin{table*}[t]
\centering
\scalebox{1.0}{
\begin{tabular}{c | c c |c| c c |c }
\hline
 \multirow{2}{*}{Model} & \multicolumn{3}{c|}{XQuAD} & \multicolumn{3}{c}{MLQA} \\
\cline{2-7} & en & hi & avg & en & hi & avg. \\ 
\hline 
mBERT & 83.9/72.9 & 58.5/43.5 & 71.2/58.2 & 80.4/67.3 & 50.3/35.2 & 65.3/51.2 \\
MuRIL &  \textbf{84.3/72.9} &  \textbf{73.9/58.3} &  \textbf{79.1/65.6} &  \textbf{80.3/67.4} &  \textbf{67.3/50.2} &  \textbf{73.8/58.8} \\
\hline
\end{tabular}}
\caption{\label{table:qa} XQuAD and MLQA (F1/EM) Results for each language. }
\end{table*}

\begin{table*}[t]
\centering
\scalebox{1.0}{
\begin{tabular}{c | c c c|c }
\hline Model & bn & en & te & avg. \\ 
\hline 
mBERT & 60.6/45.1 & \textbf{75.2/65.0} & 53.6/44.5 & 63.1/51.7\\
MuRIL & \textbf{78.0/66.4} & 74.1/64.6 & \textbf{74.0/46.9} & \textbf{75.4/59.3} \\
\hline
\end{tabular}}
\caption{\label{table:tydiqa} TyDiQA (F1/EM) Results for each language. }
\end{table*}

\begin{table*}[t]
\centering
\scalebox{1.0}{
\begin{tabular}{c | c c c c c c c| c }
\hline Model & bn-tr & hi-tr & ml-tr & mr-tr & ta-tr & te-tr & ur-tr & avg. \\ 
\hline 
mBERT & 41.8 & 25.5 & 7.5 & 8.3 & 1.0 & 8.2 & 7.3 & 14.2 \\
MuRIL & \textbf{72.9} & \textbf{69.8} & \textbf{63.4} & \textbf{68.8} & \textbf{7.0} & \textbf{53.6} & \textbf{68.4} & \textbf{57.7} \\
\hline
\end{tabular}}
\caption{\label{table:panx_tr} PANX (F1) Results for each language. }
\end{table*}

\begin{table*}[t]
\centering
\scalebox{1.0}{
\begin{tabular}{c | c c c c c | c }
\hline Model & hi-tr & mr-tr & ta-tr & te-tr & ur-tr & avg. \\ 
\hline 
mBERT & 25.0 & 33.7 & 24.0 & 36.2 & 22.1 & 28.2 \\
MuRIL & \textbf{63.1} & \textbf{67.2} & \textbf{58.4} & \textbf{65.3} & \textbf{56.5} & \textbf{62.1} \\
\hline
\end{tabular}}
\caption{\label{table:udpos_tr} UDPOS (F1) Results for each language. }
\end{table*}

\begin{table*}[t]
\centering
\scalebox{1.0}{
\begin{tabular}{c | c c| c }
\hline Model & hi-tr & ur-tr & avg. \\ 
\hline 
mBERT & 39.6 & 38.9 & 39.2 \\
MuRIL & \textbf{68.2} & \textbf{61.2} & \textbf{64.7} \\
\hline
\end{tabular}}
\caption{\label{table:xnli_tr} XNLI (Accuracy) Results for each language. }
\end{table*}

\begin{table*}[t]
\centering
\scalebox{1.0}{
\begin{tabular}{c | c c c c c c c | c }
\hline Model & bn-tr & hi-tr & ml-tr & mr-tr & ta-tr & te-tr & ur-tr & avg. \\ 
\hline 
mBERT & 1.8 & 3.0 & 2.2 & 2.4 & 2.0 & 5.1 & 2.3 & 2.7 \\
MuRIL & \textbf{8.1} & \textbf{14.9} & \textbf{10.3} & \textbf{7.2} & \textbf{11.1} & \textbf{11.5} & \textbf{13.7} & \textbf{11.0} \\
\hline
\end{tabular}}
\caption{\label{table:tatoeba_tr} Tatoeba (Accuracy) Results for each language. }
\end{table*}

\end{document}